\documentclass{article}

\usepackage{microtype}
\usepackage{graphicx}
\usepackage{subfigure}
\usepackage{booktabs} %
\usepackage{amsmath,amssymb,amsfonts}
\usepackage{calligra}
\usepackage{bbm}

\newcommand{\modelname}{Equalizer}
\newcommand{\lossnameapp}{Appearance Confusion Loss}
\newcommand{\lossnamecon}{Confident Loss}

\newcommand{\myparagraph}[1]{\vspace{3pt}\noindent{\bf #1}}

\usepackage{hyperref}

\usepackage[accepted]{icml2018}

\icmltitlerunning{Women also Snowboard: \\
Overcoming Bias in Captioning Models (Extended Abstract)}

\begin{document}

\twocolumn[
\icmltitle{Women also Snowboard: \\
Overcoming Bias in Captioning Models}

\icmlsetsymbol{equal}{*}

\begin{icmlauthorlist}
\icmlauthor{Lisa Anne Hendricks}{equal,to}
\icmlauthor{Kaylee Burns}{equal,to}
\icmlauthor{Kate Saenko}{goo}
\icmlauthor{Trevor Darrell}{to}
\icmlauthor{Anna Rohrbach}{to}
\end{icmlauthorlist}

\icmlaffiliation{to}{UC Berkeley}
\icmlaffiliation{goo}{Boston University}

\icmlcorrespondingauthor{Lisa Anne Hendricks}{lisa\_anne@berkeley.edu}

\icmlkeywords{Machine Learning, ICML}

\vskip 0.3in
]

\printAffiliationsAndNotice{\icmlEqualContribution} %

\begin{abstract}
Most machine learning methods are known to capture and exploit biases of the training data. 
While some biases are beneficial for learning, others are harmful. Specifically, image captioning models tend to exaggerate biases present in training data.  %
This can lead to incorrect captions in domains where unbiased captions are desired, or required, due to over-reliance on the learned prior and image context.  We investigate generation of gender-specific caption words (e.g. man, woman) based on the person's appearance or the image context. We introduce a new \modelname{} model that ensures equal gender probability when gender evidence is occluded in a scene and confident predictions when gender evidence is present. The resulting model is forced to look at a person rather than use contextual cues to make a gender-specific prediction. The losses that comprise our model, the \lossnameapp{} and the \lossnamecon{},  are general, and can be added to any description model in order to mitigate impacts of unwanted bias in a description dataset.  Our proposed model has lower error than prior work when describing images with people and mentioning their gender and more closely matches the ground truth ratio of sentences including women to sentences including men. %
\end{abstract}

\section{Introduction}

\begin{figure*}[t]
\centering
\includegraphics[width=\linewidth]{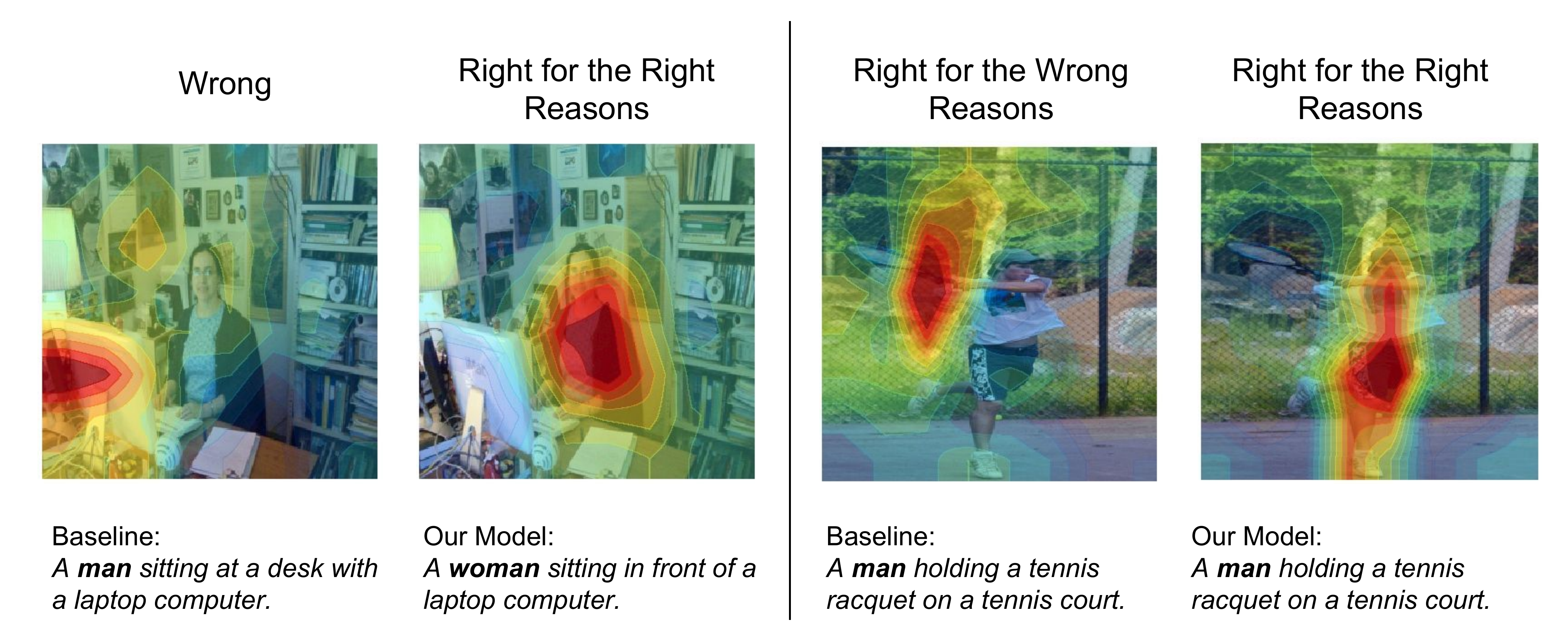}
\caption{\label{fig:teaser}  
We illustrate examples where our proposed model (\modelname{}) corrects bias in image captions. The overlaid heatmap indicates which image regions are most important for predicting the gender word.
On the left, the baseline predicts gender incorrectly, presumably because it looks at the laptop (not the person).  Even though the baseline predicts gender correctly for the example on the right, it does not look at the person when predicting gender and is thus not acceptable.  In contrast our model predicts the correct gender word and correctly considers the person when predicting gender.}
\vspace{-5mm}
\end{figure*}

Exploiting contextual cues can frequently lead to better performance on computer vision tasks~\cite{torralba2001statistical}.
For example, in visual description, predicting a ``mouse'' might be easier given that a computer is also in the image. 
However, in some cases making decisions based on context can lead to incorrect, and perhaps even offensive, predictions.
We consider one such scenario: generating captions which discuss gender.
We posit that when description models predict gendered words such as ``man'' or ``woman'', they should consider visual evidence associated with the described person, and not contextual cues like location (e.g., ``kitchen'') or other objects in a scene (e.g., ``snowboard'').
Not only is it important for description systems to avoid egregious errors (e.g., always predicting the word ``man'' in snowboarding scenes), but it is also important for predictions to be right for the right reason. For example, Figure~\ref{fig:teaser} (left) shows a case where prior work predicts the incorrect gender, while our model accurately predicts the gender by considering the correct gender evidence.
Figure~\ref{fig:teaser} (right) shows an example where both models predict the correct gender, but prior work does not look at the person when describing the image (it is right for the wrong reasons).

Bias in image captioning is particularly challenging to overcome because of the multimodal nature of the task; predicted words are not only influenced by an image, but also biased by the learned language model. 
Though recent work \cite{zhao2017men} studied bias for structured prediction tasks (e.g., semantic role labeling), image captioning was not considered.
Furthermore, the solution proposed in \cite{zhao2017men} requires access to the entire test set in order to rebalance gender predictions to reflect the distribution in the training set.
Consequently, \cite{zhao2017men} relies on the assumption that the distribution of genders is the same at both training and test time.
We consider a more realistic scenario in which captions are generated for images independent of other images in the test set.

In order to encourage description models to generate less biased captions, we introduce the \textit{\modelname{}} Model.   
Our model includes two complementary loss terms: the \textit{\lossnameapp{} (ACL)} and the \textit{\lossnamecon{} (Conf)}.
The \lossnameapp{} is based on the intuition that, given an image in which evidence of gender is absent, description models should be unable to accurately predict a gendered word.
However, it is not enough to confuse the model when gender evidence is absent; we must also encourage the model to consider gender evidence when it is present.
Our \lossnamecon{} helps to increase the model's confidence when gender is in the image.

\section{Related Work}

\myparagraph{Unwanted Dataset Bias.} 
Unwanted dataset biases (e.g., gender, ethnic biases) have been studied across a wide variety of AI domains. %
One common theme is the notion of \textit{bias amplification}, in which bias is not only learned, but amplified~\cite{bolukbasi2016man,zhao2017men}.
We find that this is of particular concern when generating visual descriptions which include gender words.

\myparagraph{Fairness.} Building AI systems which treat \textit{protected attributes} (e.g., age, gender, sexual orientation) in a fair manner is increasingly important as AI continues to become more ubiquitous~\cite{hardt2016equality}.
In machine learning literature, ``fairness'' generally considers how to ensure that systems do not use information, such as gender, in a way that disadvantages one group over another.
Our scenario is different than the general ``fairness'' setting as we are trying to \textit{predict} protected attributes.
However, a desirable caption model would be ``fair'' in the sense that it would not presume a specific gender without appropriate evidence. %

\myparagraph{Gender Classification.}  
Common description datasets frequently include captions about gender.
Thus, caption models implicitly learn to classify gender.
Our aim is to generate descriptions which discuss gender in a less biased manner.
Outside image caption generation, gender classification is generally seen as a binary task: data points are classified as either ``man'' or ``woman''.
However, AI practitioners \footnote{https://clarifai.com/blog/socially-responsible-pixels-a-look-inside-clarifais-new-demographics-recognition-model}{$^,$}\footnote{https://www.media.mit.edu/projects/gender-shades/faq}, are increasingly concerned that gender classification systems should not exclude those who do not fall in this traditional gender binary. 
In our work, we consider three gender categories: male, female, and gender neutral (e.g., person) based on visual appearance. 
We emphasize that we are not classifying biological sex or gender identity, but rather outward gender appearance, or gender expression.
\section{\modelname{}: Overcoming Bias in Description Models}

\modelname{}, is based on the following intuitions: if evidence to support a specific gender decision is not present in an image, the model should be \textit{confused} about which gender to predict (enforced by an \lossnameapp{} term), and if evidence to support a gender decision is in an image, the model should be \textit{confident} in its prediction (enforced by a \lossnamecon{} term). 
We train our model to optimize these proposed losses, as well as the standard cross entropy loss ($\mathcal{L}^{CE}$).
We use~\citet{vinyals2015show} as our base network.  %

\myparagraph{\lossnameapp{}.}  The goal of our \lossnameapp{} is to encourage the underlying description model to be \textit{confused} when making gender decisions if the input image does not contain appropriate evidence. To optimize the \lossnameapp{}, we require ground truth rationales indicating which evidence is appropriate for a particular gender decision.
We expect the resulting rationales to be masks, $M$, which are $1$ for pixels which should not contribute to a gender decision and $0$ for pixels which are appropriate to consider when determining gender.
The Hadamard product of the mask and the original image, $I \odot M$, yields a new image, $I'$, with gender information that the implementer deems appropriate for classification removed. %
Intuitively, for an image devoid of gender information, we would like the probability of predicting man or woman to be equal. 
To this end, the \lossnameapp{} enforces a fair prior by asserting that this is the case.%

To define our \lossnameapp{}, we first define a \textit{confusion} function ($\mathcal{C}$) which operates over the predicted distribution of words $p(\tilde{w}_t)$, a set of woman gender words ($\mathcal{G}_w$), and a set of man gender words ($\mathcal{G}_m$):
\begin{align*}
\mathcal{C}(\tilde{w}_t, I') = |&\sum_{g_w \in \mathcal{G}_w} p(\tilde{w}_t=g_w|w_{0:t-1}, I') - \\
&\sum_{g_m \in \mathcal{G}_m} p(\tilde{w}_t = g_m|w_{0:t-1}, I')| 
\end{align*}

Applying this definition of confusion, we can now define our \lossnameapp{} ($\mathcal{L}^{AC}$):

\vspace{-5mm}
\begin{equation}
\mathcal{L}^{AC} = \frac{1}{N} \sum_{n=0}^N \sum_{t=0}^T \mathbbm{1}(w_t \in \mathcal{G}_w \cup \mathcal{G}_m) \mathcal{C}(\tilde{w}_t, I')
\end{equation}
\vspace{-5mm}

where $\mathbbm{1}$ is an indicator variable that denotes whether or not $w_t$ is a gendered word.
For non-gendered words that correspond to images $I'$, we apply the standard cross entropy loss but do not consider loss for gendered words. %

\myparagraph{\lossnamecon{}.}
Our \lossnamecon{} encourages the probabilities for predicted gender words to be confident on images $I$ in which gender information is present. 
Given functions $\mathcal{F}^W$ and $\mathcal{F}^M$ which measure how confidently the model predicts woman and man words respectively,
we can write the \lossnamecon{} as:

\vspace{-5mm}
\begin{align*}
\mathcal{L}^{Con} = &\frac{1}{N} \sum_{n=0}^N \sum_{t=0}^T (\mathbbm{1}(w_t \in \mathcal{G}_w) \mathcal{F}^W (\tilde{w_t}, I) \\
&+ \mathbbm{1}(w_t \in \mathcal{G}_m) \mathcal{F}^M (\tilde{w_t}, I))
\end{align*}
\vspace{-5mm}

To measure the confidence of predicted gender words, we define $\mathcal{F}^W (\tilde{w_t}, I)$ as the quotient between the predicted probability for man and woman gender words and $\mathcal{F}^M (\tilde{w_t}, I)$ to be the quotient between the predicted probability for woman and man gender words.
When the model is confident of a gendered prediction (e.g., for the word ``woman''), the probability of the word ``woman'' should be considerably higher than the probability for the word ``man'', which will result in a small value for $\mathcal{F}^W$ and thus a small loss.

\myparagraph{The \modelname{} Model.}  Our final model is a linear combination of all aforementioned losses:

\vspace{-5mm}
\begin{equation}
\mathcal{L} = \alpha\mathcal{L}^{CE} + \beta\mathcal{L}^{AC} + \mu\mathcal{L}^{Con}
\end{equation}
\vspace{-5mm}

where $\alpha$, $\beta$, and $\mu$ are hyperparameters. %

\section{Experiments}

\begin{table*}[t]
\centering
\begin{tabular}{ l@{\ \ }|@{\ \ }c c@{\ \ }|@{\ \ }c c@{\ \ }|@{\ \ }c c  c }
 & \multicolumn{2} {c@{\ \ }|@{\ \ }} {MSCOCO-Bias} & \multicolumn{2} {c@{\ \ }|@{\ \ }} {MSCOCO-Confident} & \multicolumn{3} {c} {MSCOCO-Balanced} \\
Model & Error & Ratio & Error & Ratio & Error & Ratio & Pointing Game\\
  \midrule
GT     & -   &  0.466  & - & 0.548 & - & 1.000  &  - \\
  \midrule
 Baseline-FT  & 0.129   & 0.265 & 0.143 & 0.384 & 0.203 & 0.597 & 40.7 \\
 Balanced    & 0.129   & 0.270 & 0.142 & 0.393  & 0.204 & 0.610 &  37.4\\
 UpWeight    & 0.134  & 0.315 & 0.116 &  0.472 & 0.157 & 0.712  & 46.0 \\
\midrule
 \modelname{} w/o ACL  & 0.079 & 0.369 & 0.081 & 0.499 & 0.106 & 0.777 & 46.8 \\
 \modelname{} w/o Conf  & 0.098    & 0.318 & 0.116 & 0.425 & 0.165 & 0.673 &  40.5 \\
\modelname{} &  \textbf{0.070} & \textbf{0.437}  & \textbf{0.071} & \textbf{0.563} & \textbf{0.081} & \textbf{0.973} & \textbf{48.7} \\
\end{tabular}
\vspace{0.1cm}
\caption{\label{tab:accuracy-ratio} \modelname{} achieves the lowest error rate and predicts sentences with a gender ratio most similar to the corresponding ground truth captions.  \modelname{} performs best on the pointing game evaluation, which indicates if the model looks at humans when describing gender.}
\end{table*}

\myparagraph{Datasets.}  To evaluate our method, we consider the dataset used by \citet{zhao2017men} for evaluating bias amplification in structured prediction problems.
This dataset consists of images from MSCOCO~\cite{lin2014microsoft} which are labeled as ``man'' or ``woman''.  
Images are labeled as ``man''/``woman'' if at least one description included the word ``man''/``woman'' and no descriptions include the word ``woman''/``man''.
We call this dataset \textbf{MSCOCO-Bias}.
Each image in MSCOCO is provided with five ground truth captions, and sometimes not all captions mention a gender word.
We therefore construct an \textbf{MSCOCO-Confident} set in which only images in which at least four annotations include a gender word are included.
Finally, in addition to the MSCOCO-Bias set, we also evaluate on a set where we purposely change the ratio of men and women.
We select 500 images from MSCOCO-Confident which include the word ``woman'' and 500 images which include the word ``man'' to create the \textbf{MSOCO-Balanced} set.

\myparagraph{Metrics.}  The first metric we consider is \textbf{error rate}, or the number of man/woman misclassifications. %
Second we consider the \textbf{gender ratio} of sentences which belong to a ``woman'' set to sentences which belong to a ``man'' set. %
Finally, to measure if our models are \textbf{right for the right reasons} we consider the pointing game \cite{zhang2016top} evaluation  on the MSCOCO-Balanced dataset. We create visual explanations using Gradient-weighted Class Activation Mapping (Grad-CAM) system~\cite{selvaraju2016grad}.  %

\myparagraph{Baselines and Ablations}
We call our first baseline \textbf{Baseline-FT}. We fine-tune the Show and Tell model~\cite{vinyals2015show} through the LSTM and convolutional networks using the standard cross-entropy loss on our target dataset. %
Additionally, we consider a \textbf{balanced} baseline where we re-balance the data distribution at training time to account for the larger number of men instances in the training data. 
We also consider an \textbf{upweight} baseline where we upweight the loss value for gender words in the standard cross entropy loss to increase the penalty for a misclassification and the reward for a correct classification. %
To isolate the impact of the two loss terms in \modelname{}, we report results with only the Appearance Confusion Loss (\modelname{} w/o Conf) and only the Confidence Loss (\modelname{} w/o ACL).

\textbf{Results.}
In Table~\ref{tab:accuracy-ratio} we analyze the error rates when describing men and women by looking at the captions generated for the MSCOCO-Bias, MSCOCO-Confident and MSCOCO-Balance set.
Our full model improves upon \modelname{} w/o ACL and \modelname{} w/o Conf on all three datasets.
When comparing \modelname{} to baselines, we see the largest performance difference on the MSCOCO-Balanced dataset.
This is in part because our model does a particularly good job of decreasing error on the minority class (woman).

In addition to error rates, we can consider the accuracy breakdown for specific genders.  Generated captions can include male, female, or gender neutral words to describe people.
The Basline-FT model predicts ``man'' correctly with 86.2\% accuracy and ``woman'' correctly with 62.8\% accuracy on the MSCOCO-Confident set. 
In contrast, Equalizer predicts ``man'' correctly with 75.9\% accuracy and ``woman'' correctly with 74.6\% accuracy.
Interestingly, the Equalizer model has similar accuracy for both man and woman words.
Though the recall for man is not as high using Equalizer (75.9\% vs. 86.2\%), Equalizer less frequently misclassifies a man as a woman (5.6\% vs 5.7\%).  
This is in part because the Equalizer model is more cautious and predicts gender neutral words more frequently (18.5\% vs. 8.1\%).

We also report the ratio of captions which include only female words to captions which include only male words. Ideally, we would like this ratio to match the respective ratio for the ground truth captions.
Impressively, \modelname{} achieves the closest ratio to ground truth for all datasets. %

Finally, we evaluate whether models are ``right for the right reason'' using the pointing game on the MSCOCO-Balanced dataset.
Ground truth masks are obtained from the MS~COCO ground-truth person segmentations.
For an easier and fair comparison we provide all models with ground-truth captions. We then generate heat maps for ``man'' and ``woman'' words present in these captions using Grad-CAM. 
\modelname{} consistently performs the best on this evaluation.
Figure~\ref{fig:teaser} includes qualitative examples.

\myparagraph{Discussion.}  We present \modelname{} model which produces less biased captions when describing gender.  
Instead of relying on balanced datasets, which can be hard to create when considering natural language, \modelname{} includes two novel loss functions which encourage the generation of fair captions.
We hope our work encourages others to consider bias in captioning models.

\bibliography{egbib}
\bibliographystyle{icml2018}

\end{document}